\documentclass{article}
\usepackage{spconf,amsmath,graphicx}
\usepackage{multirow}
\usepackage{booktabs}
\usepackage{algorithm}  
\usepackage{algpseudocode}  
\usepackage{amsthm,amsmath,amssymb}
\usepackage{mathrsfs}
\usepackage{color}
\usepackage{xcolor}

\title{Mixed Precision Quantization of \\
Transformer Language Models for Speech Recognition}
\name{Junhao Xu, Shoukang Hu, Jianwei Yu,  Xunying Liu, Helen Meng}
\address{The Chinese University of Hong Kong, Hong Kong SAR, China}
\begin{document}
\ninept
\maketitle
\vspace{-2ex}
\begin{abstract}
State-of-the-art neural language models represented by Transformers are becoming increasingly complex and expensive for practical applications. Low-bit deep neural network quantization techniques provides a powerful solution to dramatically reduce their model size. Current low-bit quantization methods are based on uniform precision and fail to account for the varying performance sensitivity at different parts of the system to quantization errors. To this end, novel mixed precision DNN quantization methods are proposed in this paper. The optimal local precision settings are automatically learned using two techniques. The first is based on a quantization sensitivity metric in the form of Hessian trace weighted quantization perturbation. The second is based on mixed precision Transformer architecture search. Alternating direction methods of multipliers (ADMM) are used to efficiently train mixed precision quantized DNN systems. Experiments conducted on Penn Treebank (PTB) and a Switchboard corpus trained LF-MMI TDNN system suggest the proposed mixed precision Transformer quantization techniques achieved model size compression ratios of up to 16 times over the full precision baseline with no recognition performance degradation. When being used to compress a larger full precision Transformer LM with more layers, overall word error rate (WER) reductions up to 1.7\% absolute (18\% relative) were obtained.
\end{abstract}
\begin{keywords}
Language models, Speech recognition, Transformer, Quantization, ADMM
\end{keywords}
\vspace{-1.5ex}
\section{Introduction}
\label{sec:intro}
\vspace{-0.5ex}
Deep Transformer models in recent years have defined state-of-the-art language modelling 
performance across a range of applications including automatic speech recognition (ASR). 
The Transformer model architecture features a deep stacking of multiple self-attention layers~\cite{jcheng2016, zlin2017, parikh2016}
 with residual connections~\cite{khe2016} and layer normalization~\cite{jlba2016}
Additional positional encoding layers~\cite{vaswani2017, Gehring2017} can be used to further augment the self-attention 
layers with sequence order information. Performance improvements over the conventional long 
short-term memory recurrent neural network (LSTM-RNN) language models have been widely 
reported~\cite{Kazuki2019, zeyer2019comparison}. However, the deeper architecture design of Transformers not only leads to a 
large increase in overall system complexity, memory footprint and computational cost when 
operating on the cloud, but also creates difficulty when deploying them on edge devices to 
enhance privacy and reduce latency, in common with many other computational intensive deep 
learning applications that are currently facing similar challenges.  

To this end, one powerful solution recently drawing increasing interest in the machine learning and speech technology community is to use low-bit deep neural network (DNN) quantization techniques~\cite{Asanovic2019, Soudry2014}. By replacing floating point weights with low precision values, for example, binary numbers, quantization can dramatically reduce the model size without modifying the network architecture~\cite{Rastegari2016, jwu2016, azhou2017}. Further model size reduction can be obtained when low-precision quantization is used in combination with neural architecture search (NAS) methods, for example, in the SqueezeNet system~\cite{iandola2016squeezenet}. In contrast to the extensive research works on low-bit quantization methods conducted on computer vision tasks~\cite{zdong2019, kwang2019}, only limited previous research in this direction has been conducted in the context of language modelling~\cite{bengio2003neural} and ASR systems.

Two issues are associated with current low-bit DNN quantization methods. First, these quantization approaches are predominantly based on uniform precision, where an identical bit-width is applied to all weight parameters for quantization. This fails to account for the varying performance sensitivity at different parts of the system to quantization errors. In practice, this often leads to large performance degradation against full precision models. Second, gradient descent methods and back-propagation (BP) algorithm cannot be directly applied in quantized model training when the weights are restricted to discrete values. Existing methods of training low-bit quantized DNNs often use a modified BP algorithm~\cite{Hubara2017, Courbariaux2015}, where low precision quantized parameters were first used in the forward pass to compute the error loss before full precision parameters are used in the backward pass to propagate the gradients for model update. However, the direct use and estimation of quantized weights in these methods leads to very slow convergence in training, while the performance gap against full precision models remains. 

In order to address these issues, novel mixed precision DNN quantization methods are proposed in this paper to address this problem by applying locally variable bit-widths to individual components of the system. These methods are becoming well supported by the recent development of mixed precision DNN acceleration hardware that allow multiple locally set precision settings to be used~\cite{zdong20192}. The resulting flexibility can provide a better trade-off between compression ratio and accuracy performance target.  The optimal local precision settings are automatically learned using two techniques. The first is based on a quantization sensitivity metric in the form of Hessian trace weighted quantization perturbation. It can be efficiently computed using Hessian-free approaches. The second is based on mixed precision Transformer architecture search. 

In order to overcome the difficulty in using gradient descent methods to directly estimate DNNs of discrete quantized weights, alternating direction methods of multipliers (ADMM) are proposed to efficiently train mixed precision quantized DNN systems. Experiments conducted on multiple tasks: Penn Treebank (PTB), Switchboard (SWBD) suggest the proposed mixed precision Transformer LM quantization techniques achieved a model size compression ratio of up to 16 times over the full precision baseline with no recognition performance degradation. Moreover, by applying quantization to a more complex Transformer LM with more layers, we can get overall WER reduction up to 1.7\% absolute.

The main contributions of this paper are summarized as following. First, to the best of our knowledge, this paper is the first work to apply mixed precision quantization methods to Transformer language models. In contrast, previous researches on low-bit quantization focused on convolutional neural networks (CNNs)~\cite{yqian2019} and LSTM-RNN LMs~\cite{kyu2018}, where expert designed special partially quantized linear layers containing binary weight matrices, full precision bias and additional scaling parameters were used mitigate the performance degradation due to uniform precision quantization.
 %Such manually designed mixed precision quantization configurations cannot be automatically learned as allowed by the methods presented in this paper. 
 
The rest of the paper is organized as follows. Transformer LMs are reviewed in section 2. A general neural network quantization scheme and uniform quantization are presented in section 3. Section 4 presents our mixed precision quantization methods in details. Experiments and results are shown in section 5. Finally, conclusions and future work are discussed in section 6. 
\vspace{-1.5ex}
\section{Transformer LMs}
\vspace{-0.5ex}
\label{sec:transformerlm}
The Transformer model architecture considered in this paper feature a deep stacking of multihead attention followed by feedforward layers. Residual connections and layer normalization are also inserted between them as in the top part of figure 1. The $l$-th Transformer layer transforms the 
input $\bf{x}^{l-1}$ at $t$ time step as follows:
\begin{align}
    {\bf q}_{t}^{l}, {\bf k}_{t}^{l}, {\bf v}_{t}^{l}&=\mathbf{Q}{\bf x}_{t}^{l-1},\mathbf{K}{\bf x}_{t}^{l-1},\mathbf{V}{\bf x}_{t}^{l-1} \\
    {\bf h}_{t}^{l} &= ({\bf h}_{t-1}^{l}, ({\bf k}_{t}^{l},{\bf v}_{t}^{l})) \\
    {\bf y}_{t}^{l} &= \mathbf{W}_h^{l}\text{SelfAttention}({\bf h}_{t}^{l}, {\bf q}_{t}^{l}) + {\bf x}_{t}^{l-1} \\
    {\bf z}_{t}^{l} &= \text{LayerNorm}({\bf y}_{t}^{l})
\end{align}
where $\bf{Q}, \bf{K}, \bf{V}$ are the query, key, value projection matrices. $\bf{h}^l$ stores the history information in the l-th self-attention layer. $(, )$ denotes vector concatenation operation. $\text{SelfAttention}(\cdot)$ is the scaled multi-head dot product self-attention machanism and $\bf{W}_h^l$is the projection matrix. $\text{LayerNorm}(\cdot)$ is the layer normalization. 

The feed-forward layer at time step $t$ is:
\begin{align}
    {\bf s}_{t}^{l} &= \mathbf{W}_2^{l}\text{GELU}(\mathbf{W}_1^l{\bf z}_{t}^{l}+\bf{b}_1^l) + \bf{b}_2^l + {\bf z}_{t}^{l} \\
    {\bf x}_{t}^{l} &= \text{LayerNorm}({\bf s}_{t}^{l})
\end{align}
where $\bf{W}_1^l$ and $\bf{W}_2^l$ are the weight matrices and $\bf{b}_1^l$ and $\bf{b}_2^l$ are the corresponding bias. $\text{GELU}(\cdot)$ represents the Gaussian error linear unit~\cite{Hendrycks2018}. In addition, we also use positional embedding layer in the transformer LMs.

\vspace{-1.5ex}
\section{Neural Network Quantization}
\vspace{-0.5ex}
\label{sec:nnquantization}
For a standard n-bit quantization problem of neural networks, we consider a full precision weight parameter $\theta$ and find its closest discrete approximation from the following quantization table $q \in \{0, \pm1,\pm2, \dots, \pm (2^{n-1}-1)\}$ as
\begin{equation}
 f(\theta) = \arg\min \limits_{q} |\theta - q|
 \end{equation}
1 bit is reserved to denote sign bit. With further simplification, extremely low bit quantization, for example, binarization $\{1, -1\}$~\cite{leng2018extremely, rastegari2016xnor} and ternary $\{-1, 0, 1\}$~\cite{li2016ternary}, can be produced.

Applying quantization to all weight matrices in the model, we can use a more general format in equation (7) to represent the quantization for each parameter. Let $\theta^{(l)}_{i}$ be the $i^{th}$ parameter within any of the $l^{th}$ weight cluster, for example, all weight parameters of the same layer,

 \begin{equation}
  \begin{split}
  \label{eq:quanMap}
  f(\theta^{(l)}_i) = \arg\min \limits_{Q_i^{(l)}} |\theta^{(l)}_i - Q_i^{(l)}|
\vspace{-1cm} 
\end{split}
\end{equation}

The locally shared $l^{th}$ quantization table is given by

\begin{equation}
\label{equ:31}
  {Q}_i^{(l)}\in\{0, \alpha^{(l)},\dots, \alpha^{(l)} (2^{n-1}-1)\}
\end{equation}
where $\alpha^{(l)}$ is a full precision scaling factor used to adjust the dynamic range of all the unquantized weights in the cluster. It is shared locally among weight parameters clusters. A special case, when the local quantization table in equation (8) is shared across all the layers, leads to the traditional uniform precision quantization approach. The only remaining factor affecting the system performance is the bit length $\#bit$ which is also globally set to be $1, 2, 4, 8$ etc.

%\vfill\pagebreak
\vspace{-1.5ex}
\section{Mixed Precision Transformer Quantization}
\vspace{-0.5ex}
\label{sec:mixedquant} 
This section presents three mixed precision based Transformer LM quantization approaches.

\vspace{-1.5ex}
\subsection{ADMM Based Mixed Precision Quantization}
One major challenge faced by both uniform and mixed precision quantization is that the gradient descent methods and back-propogation (BP) algorithm can not be directly used when weights are quantized to discrete values. To this end, mixed precision BP was proposed later in~\cite{Courbariaux2015} where low precision binarized parameters were first used in the forward pass to compute the error loss before full precision parameters are used in the backward pass to propagate the gradients. However, directly training quantized system using mixed precision BP leads to very slow convergence and the performance gap between full precision and quantized systems remains large. An alternative solution to this problem is to reformulate quantization as a constrained optimization problem implemented solved by the alternating direction methods of multipliers (ADMM)~\cite{boydS2011}. It was initially used to in~\cite{xu2018alternating} learn the global quantization table in equation (8) where $\alpha$ is shared among all the parameters. 

In order to account for the locally varying performance sensitivity, ADMM was used in our earlier research~\cite{junhao2019} to learn the local quantization tables in equation (8). This allows ADMM to provide a form of mixed precision quantization. However, the optimum local quantization precision settings cannot be learned by ADMM and must be manually set. These will be automatically learned in the following two approaches of Sections 4.2 and 4.3.

\vspace{-1.5ex}
\subsection{Minimum Sensitivity Based Mixed Precision Quantization}

\begin{figure}[htbp]
%    \begin{center}
    \includegraphics[scale=0.26]{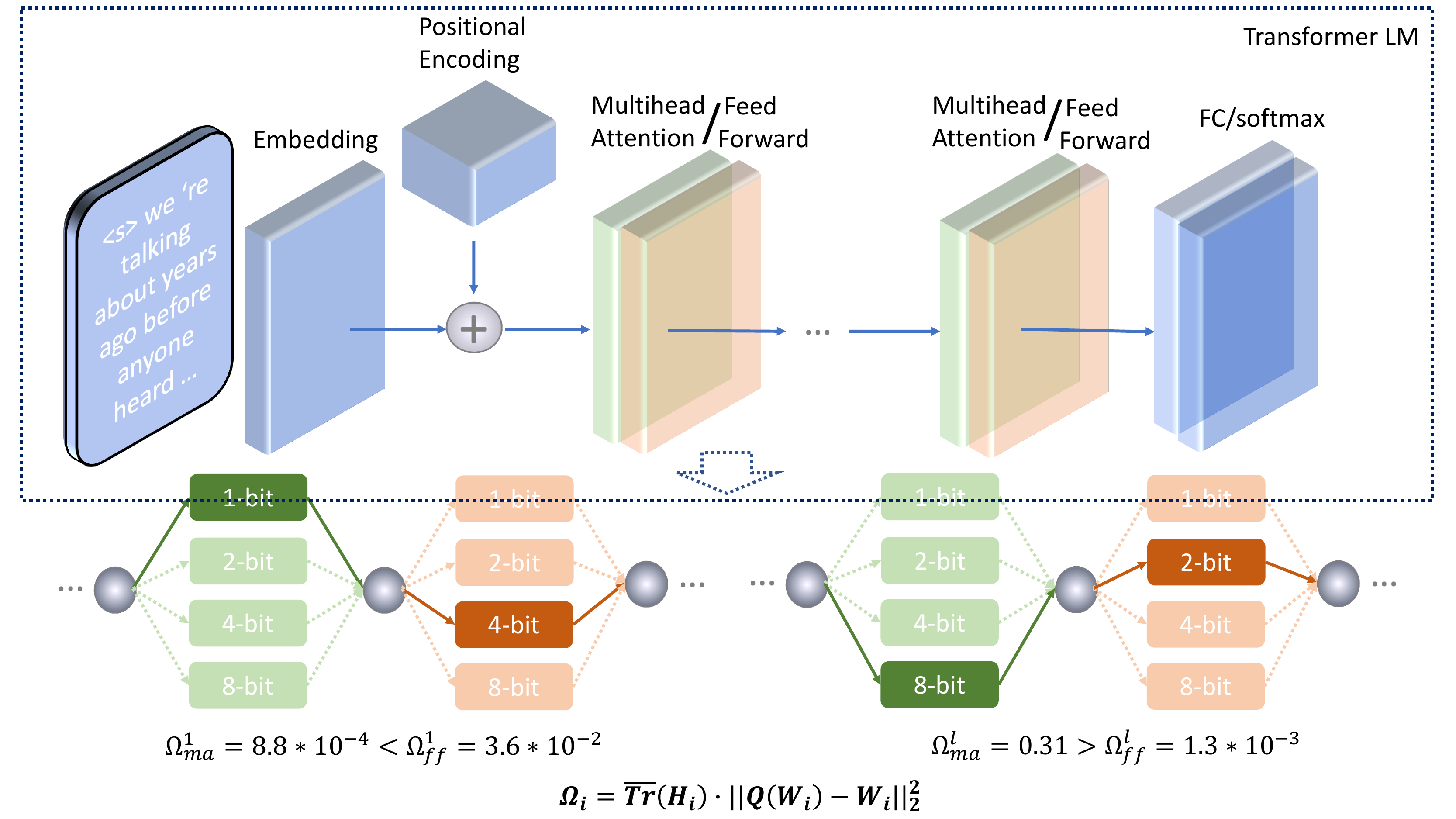}
     \vspace{-1,5em}
    \caption{\emph{An example of auto-configured mixed precision quantization of a transformer LM using a minimum performance sensitivity measure. For the first transformer module positioned right after the embedding and position encoding layer, its multi-head attention layer (green) is uses binary quantization while its feed forward layer (orange) uses 4-bit quantization precision, as determinted by the Hessian-trace weighted quantization sensitivity measure.}}
    \label{fig:per}
        \vspace{-2em}
 %   \end{center}
\end{figure}
Assuming the parameters of a neural network is twice differentiable and converged to a local optimum, it was proved in~\cite{zdong2019} that the expected performance loss, when using a given quantization precision, is expressed in the form of Hessian trace weighted squared quantization error. In simple terms, for each cluster of weight parameters, given the same amount of weight perturbation resulted from quantization, the smaller the associated Hessian matrix trace, the lower the performance sensitivity to quantization. 

\noindent For any quantization $\bf{Q}(\cdot)$ being applied to the network parameters $\bf{W}$, the total performance sensitivity can be represented by the following sum of Hessian trace and squared quantization perturbation error. 
\begin{equation}
    \Omega = \sum_{i=1}^L\Omega_i = \sum_{i=1}^L\Bar{Tr}(\bf{H}_i)\cdot||Q(\bf{W}_i)-\bf{W}_i||_2^2
\end{equation}

Given a target average quantization precision, the local quantization bit widths used in each layer should be selected such that the above total performance sensitivity is minimized. In practice, this requires transformer LMs using uniform precision, for example 1-bit, 2-bit, 4-bit and 8-bit be separately trained off-line first via ADMM optimization in Section 4.1. The performance sensitivity in equation (9) can then be computed locally for each layer using each quantization choices before taking the sum.

For larger transformer LMs containing 
%hundreds of 
millions of parameters, and many large deep neural networks in general, directly computing the Hessian matrix and its trace is infeasible. In order to handle this
% issue
, an efficient stochastic linear algebra approach based on the Huchinson’s Algorithm~\cite{avron2011randomized} is used to approximate the Hessian trace,
\vspace{-1ex} 
\begin{equation}
Tr(\bf{H})\approx \frac{1}{m}\sum_{i=1}^m \bf{z}_i^T\bf{H}\bf{z}_i 
   \end{equation}
 
   where the expensive matrix multiplication between $\bf{H}$ and $\bf{z}_i$ in the above approximation can be avoided, and efficiently computed using Hessian-free approaches~\cite{zdong2019}. $\bf{z}_i$ is a random vector sampled from a Gaussian Distribution $\mathcal{N}(\bf{0}, \bf{1})$.
 
 An example application of the minimum performance sensitivity based mixed precision quantization of two layers within a Transformer LM is shown in figure 1 (residual connection and normallzation are omitted for brevity). 
 
 \vspace{-1.5ex}
\subsection{Architecture Search Based Mixed Precision Quantization}

\begin{figure}[htbp]
%    \begin{center}
    \includegraphics[scale=0.26]{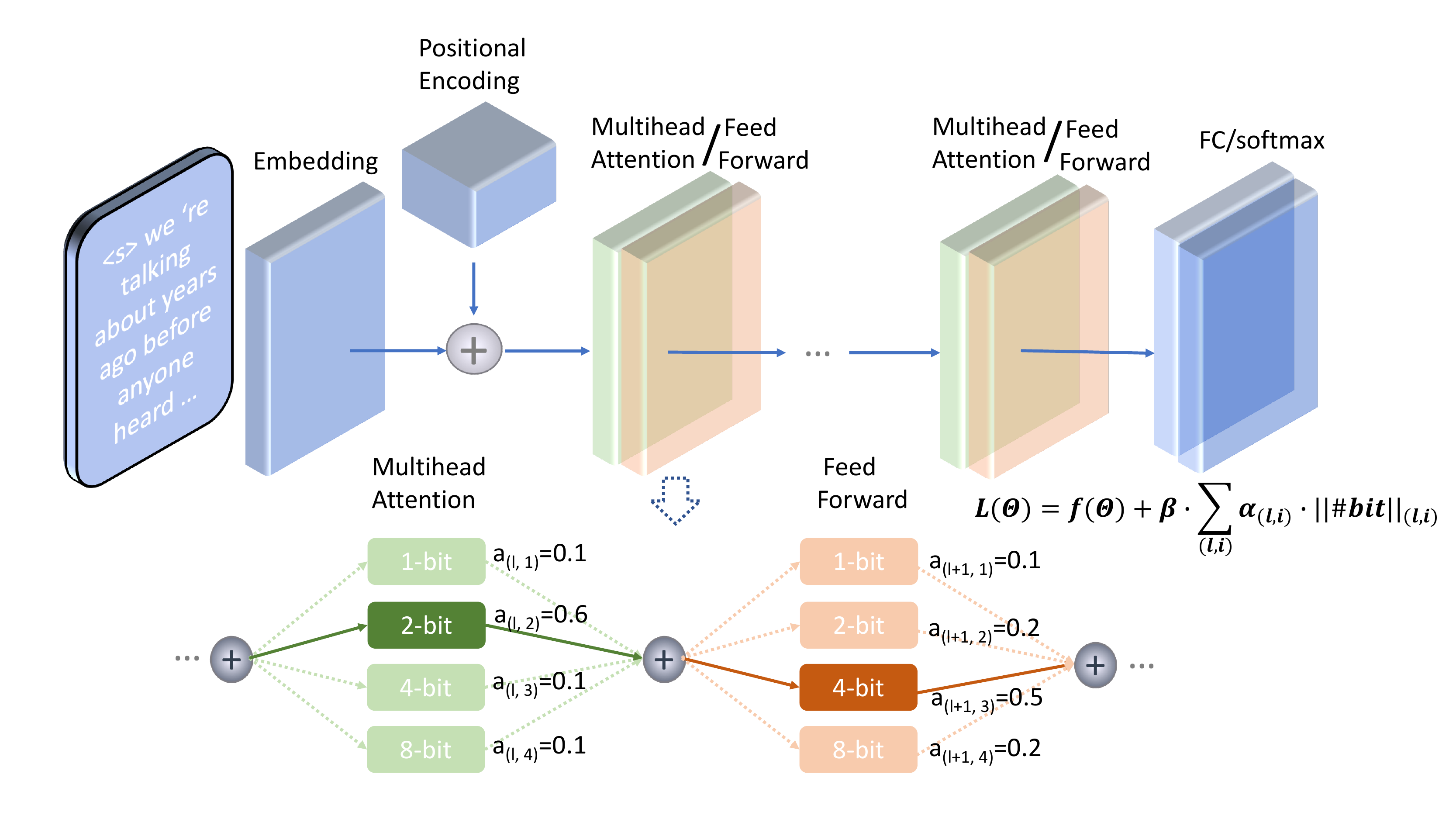}
     \vspace{-1,5em}
    \caption{\emph{An example of auto-configured mixed precision quantization of a transformer LM using mixed precision architecture search. For the first transformer module, its multi-head attention layer is uses 2-bit quantization (green) given the associated selection weight of 0.6 while its feed forward layer uses 4-bit quantization precision (orange) given the associated selection weight of 0.5, as the 1-best choice selected from the mixed precision NAS super-network.}}
    \label{fig:per}
        \vspace{-2em}
 %   \end{center}
\end{figure}

An alternative solution to automatically determine the suitable local quantization precision settings is to use mixed precision neural architecture search (NAS)~\cite{liu2018progressive}. Inside a NAS super-network containing all possible Transformer architectures with varying precision bit widths, the differentiable architecture weights~\cite{liu2018darts} associated different precision settings can be automatically learned inside the super-network together with the normal Transformer parameters. 

Instead of selecting over heterogeneous neural building structures as considered in conventional NAS applications, 
%for example, convolutions and context pooling, 
now for transformer LM quantization purposes, different neural building blocks, for example, Transformer modules of different bit-widths are considered. This major difference requires the associated mixed precision quantization super-network to be specially designed. 
Such super-network is constructed by first separately training transformer LMs using uniform precision, for example 1-bit, 2-bit, 4-bit and 8-bit, using ADMM optimization, before connecting these uniform precision quantized Transformer LMs at each layer, where the system specific activation outputs are linearly combined using a set of quantization precision selection weights (\{$\alpha_{(l,i)}$\} in equation (12)). An example of such mixed precision Transformer super-network is shown in Figure 3. 

In order to avoid the trivial selection of the longest, most generous quantization bit width, these precision selection weights learning can be further constrained by a model complexity penalty in terms of the number of bits retained after quantization. \footnote{The square root of \#bit provides a smoothing effect on the precision system complexity penalty term to avoid over-penalizing high precision settings}
\vspace{-1ex} 
\begin{equation}
    L(\boldsymbol{\theta})= f(\boldsymbol{\theta}) + \beta\sum_{(i, l)} \boldsymbol{\alpha}_{(i,l)}\cdot\sqrt{||\#bit||_{(i, l)}} 
       \vspace{-1em}
\end{equation} 

where $f(\boldsymbol{\theta})$ is the standard cross-entropy loss.

\vspace{-1.5ex}
\section{Experiments}
\vspace{-0.5ex}
In order to evaluate the performance of mixed precision quantized Transformer LMs, an initial set of experiments on the Penn Treebank (PTB) corpus are first presented in Section 5.1. The main set of experiments conducted on a Switchboard (SWBD) corpus are presented in Section 5.2. All mixture precision quantized Transformer LMs use layer level locally shared quantization tables with varying precision settings that are either manually in case of ADMM, or learned by minimum performance sensitivity (\emph{MinSen}) and mixed precision neural architecture search (MPNAS) of Sections 4.2 and 4.3. Statistical significance test was conducted at level $\alpha=0.05$ based on matched pairs sentence\-segment word error (MAPSSWE) for recognition performance analysis.

  \begin{table*}[h]
    \caption{\emph{Performance of the baseline full precision, uniform precision quantized and layer level mixed precision quantized Transformer LMs with local precision set either manually in ADMM, or automatically using MinSen/MPNAS of Sections 4.2 \& 4.3  on Switchboard NIST Hub5’00, RT02 and RT03. Evaluation time is computed over rescoring all the N-best lists.}}
    \label{tab:swbd}
    \centering
 \resizebox{175mm}{28mm}{\color{darkgray}{
    \begin{tabular}{c|c|c|c|c|c|ccc|cc|cc|c|c|c} 
        \toprule
            \multirow{2}{*}{\textbf{models}}& \textbf{quant.}& \textbf{quant.}&\multirow{2}{*}{\textbf{\#bit}} &\multirow{2}{*}{\textbf{PPL}}& +4gram & \multicolumn{3}{c|}{\textbf{rt02}}& \multicolumn{2}{c|}{\textbf{rt03}} & \multicolumn{2}{c|}{\textbf{eval2000}} & \textbf{model} & \textbf{comp.} & \textbf{evaluation}  \\  
            &\textbf{precision}&\textbf{method} & & &\textbf{PPL} & swbd1 & swbd2 & swbd3 & fsh & swbd & swbd. & callhm.& \textbf{size(MB)} & \textbf{ratio} & \textbf{time(s)}  \\ \midrule
            % lstm& \multirow{2}{*}{-} & \multirow{2}{*}{-}& 32 & 89.3 & & & & & & - &  \\ \cline{1-1}\cline{4-13} 
            1 & & & 32 & 41.24 &41.08 & 9.5 & 12.9 & 17.4 & 10.4 & 17.3& 7.8 & 15.6 & 106 & - & 13.19 \\ \hline
            2 & & \multirow{4}{*}{-}& 1 & 48.26 & 47.95 & 10.5 & 13.6  & 18.6 & 11.3 & 18.5 & 8.2 & 16.2 & 3.6 & 30.5 & 4.76 \\ 
            3 & \emph{uniform} & & \boxed{2} & 44.62 & 43.28 & 10.1 & 13.4 & 18.0 & 10.9 & 18.2 &8.1 &15.9 & 7.9 & 13.4 & 6.43 \\ 
            4 & precision & & 4 & 43.83 & 42.97 & 9.7 & 13.2 & 17.6 & 10.7 & 17.8 & 8.0 & 15.7 & 14.1 & 7.5 & 6.89 \\ 
            5 & & & 8 & 43.72 & 41.36 & 9.6 & 13.0 & 17.5 & 10.6 & 17.4& 7.9 & 15.8 & 27.2& 3.9 & 7.12 \\ \hline
            6 &  & ADMM & 1 & 47.26 & 46.10 & 10.3 & 13.5 & 18.2 & 11.1 & 18.3 & 8.1 & 16.1 &3.6 & 30.5 & 4.76 \\ 
            7 & & (\emph{Multi.} & \boxed{2}& 42.62 & 42.32 & 9.9 & 13.3 & 17.9 & 10.8 & 17.9 & 8.0 & 15.7 &7.9 & 13.4 & 6.43 \\ 
            8 & \emph{mixed}& \emph{Eql.} & 4 & 42.83 & 41.66 & 9.6 & 13.1 & 17.6 & 10.5 & 17.4 & 7.9 &15.7 &14.1 & 7.5 & 6.89\\ 
            9 & precision& \emph{Bit-width)} & 8 & 42.72 & 41.21 & 9.6 & 13.0 & 17.5 & 10.4 & 17.4 & 7.8 & 15.8 &27.2& 3.9 & 7.12  \\ \cline{3-16}\cline{1-1}
            10 & & MinSen & \boxed{\textbf{1.9}} & \textbf{42.39} & \textbf{41.52} & \textbf{9.7} & \textbf{13.0} &
            \textbf{17.6} &
            \textbf{10.6} &
            \textbf{17.5} & \textbf{7.9} & \textbf{15.7} & \textbf{8.0} & \textbf{13.25} & \textbf{6.58} \\ \cline{3-16}\cline{1-1}
            11 & & NAS. & \boxed{2.5} & 42.75 & 41.96 & 9.9 & 13.2 & 17.7 & 10.8 & 17.8 & 7.9 & 15.8 & 9.1 & 11.65 & 6.80 \\ \bottomrule
        \end{tabular}}}
            \vspace{-1.5em}
  \end{table*}

\begin{table}[h]
    \vspace{-0.5em}
    \caption{\emph{Perplexity (PPL), quantization bit length \#bit, model size and compression ratio  of baseline full precision Transformer (\emph{trans.}), uniform and mixed precision quantized Transformer using manual setting (\emph{ADMM}), performance sensitivity based quantization (\emph{MinSen}) and architecture search based quantization (\emph{NAS})}}
    \label{tab:swbd}
    % \centering
  \resizebox{85mm}{25mm}{
  \color{darkgray}{
    \begin{tabular}{c|c|c|c|c|c|c} 
        \toprule
          \multirow{2}{*}{\textbf{models}} & \textbf{quant.} & \textbf{quant.} & \multirow{2}{*}{\textbf{\#bit}} &  \multirow{2}{*}{\textbf{PPL}} & \textbf{Model.} & \textbf{Comp.}\\ 
             & \textbf{prec.}&\textbf{meth.}& & & \textbf{Size} & \textbf{Ratio}\\ \midrule
        %    lstm & \multirow{2}{*}{-} &\multirow{2}{*}{-}& 32 & 85.3  \\ \cline{1-1} \cline{4-5}
          1 & & &  32  & 55.26 & 66 & - \\ \hline
          2 &  & \multirow{4}{*}{-}&1 & 82.10& 2.3 & 28.7 \\
        %   11  & & NAS. & \textcolor{red}{2.2} & \textcolor{red}{58.23}  & 4.8 & 13.8 \\ \bottomrule
          3  & \emph{uni.} & & \boxed{2}
          & 58.94 &  4.6 & 14.3 \\
          4  & prec.& & 4  & 56.86 & 9.4 & 7.0  \\  
          5  & & & 8  & 56.80 & 17.0 & 3.9 \\ \hline
          6  &  & ADMM  &1  & 65.41 &2.3 & 28.7  \\ 
            7  &  & \emph{(Multi.}  & \boxed{2} & 58.06 & 4.6 & 14.3\\
            % 7  &  &ADMM  & \boxed{\textcolor{red}{2}} & \textcolor{red}{58.06} & 4.6 & 14.3\\
8          & \emph{mixed} & \emph{Eql.}& 4 &  56.84 &9.4 & 7.0 \\
  9    & prec. & \emph{Bit-width)} & 8 & 56.75 & 17.0 & 3.9 \\ \cline{1-1}\cline{3-7}
    10& & MinSen & \textbf{\boxed{2.0}} &  \textbf{{56.82}} & \textbf{{6.5}} & \textbf{{10.2}} \\ \cline{3-7}\cline{1-1}
    11  & & NAS. & \boxed{2.2} & 58.23  & 4.8 & 13.8 \\ \bottomrule
        %   11  & & NAS. & \boxed{\textcolor{red}{2.2}} & \textcolor{red}{58.23}  & 4.8 & 13.8 \\ \bottomrule
        \end{tabular}}}
         \vspace{-2em}
  \end{table} 

\vspace{-1.5ex}
\subsection{Experiments on Penn Treebank Corpus}
The PTB corpus uses a 10K word vocabulary. 930K words of text data were used for training. 74K and 82K words of development and test data sets were used. 

There are several trend can be found in Table 2, given the same quantization precision, for example, at approximately 2 bits, all the mixed precision quantized models ADMM (line 7), MinSen (line 10) and MPNAS (line 11), consistently outperform the 2-bit uniform quantization model in line 3.  Second, among all the mixed precision quantization methods, the lowest PPL of $56.82$ is obtained using MinSen with a quantization ratio of 10.2 times.

  \vspace{-1em}
\subsection{Experiments on Conversational Telephone Speech}

% \begin{strip}

% \end{strip}
The Switchboard I telephone speech corpus we use consists of approximately 300 hours of audio data released by LDC (LDC97S62). The baseline GMM-HMM system with 6008 tied tri-phone states was trained based on 40-dimensional Mel-frequency cepstral coefficients (MFCCs) to generate alignments for the neural network training. LF-MMI trained TDNN\cite{daniel} acoustic models with data augmentation and i-Verctor adaptation~\cite{dehak2011front} were used. Various Transformer LMs trained on the Switchboard and Fisher transcripts (LDC2004T19, LDC2005T19) was used to rescore the 4-gram LM\footnote{the same 4\-gram LM was used in both the initial lattice and N-best list generation stage, and subsequent N-best rescoring.} produced N-best lists ($N=20$). Their performance are shown in Table 1. 

Similar trends can be found in Table 1. First, given the same quantization precision, for example at approximately 2 bits, all the mixed precision quantized systems (ADMM, MinSen and MPNAS) outperform the equivalent 2-bit uniform quantized systems in line 3. Second, among the three mixed precision quantization approaches, auto-configured quantization by MinSen or MPNAS outperform the manual ADMM quantization using a comparable bit width of 2. In particular, the 1.9 bit quantized MinSen model (line 10) outperforms the comparable uniform precision (line 3) and manual ADMM quantization (line 7) by 0.4 to 0.7 and 0.3 to 0.4 absolution on rt02 and rt03 in a statistically significant manner. Finally, the evaluation time is also halved against full precision baseline.

The advantages of mixed precision quantization are further demonstrated when being used to compress a 16-layer larger Trans- former LM (line2 in table 3), The 4-bit ADMM quantized model produced the best performance, with a statistically significant WER reductions of 1.7% absolute over the over-fitting 16 layer full preci- sion Transformer on the SWBD subset of Hub5’00.

\vspace{-2ex}
\section{Conclusions} 

This paper presents a set of novel mixed precision based Transformer LM quantization techniques for the locally varying performance sensitivity to the use of low-bit precision during model compression. The optimal local precision settings are automatically learned by either minimizing the performance sensitivity, or mixed precision NAS. Experimental results conducted on state-of-the-art speech recognition tasks suggest the proposed mixed precision quantization methods outperform uniform precision based quantization, and can produce large model size compression ratios of up to 16 times over the full precision baseline with no performance degradation. Future research will focus on improving mixed precision quantization methods and their application to other ASR system components.
\begin{table}[H]
    \vspace{-1.5em}
    % \caption{Perplexity (PPL) and word error rate (WER) performance and compression ratio of quantized LSTM RNNLMs on SWBD dataset: full precision baseline with no quantization (STD), binarized model w/o  partially quantized linear layers (Bin+Lin or Bin) trained using 50 or 400 epochs, and ADMM quantized models with a layer, node or no tying of quantization tables of varying \#bits. }
    \caption{\emph{Performance of the baseline 6 or 16 layer full precision, and mixed precision quantized Transformer LMs with local precision set either manually in ADMM, or automatically using MinSen/MPNAS of Sections 4.2 \& 4.3  on Switchboard NIST Hub5’00, RT02 and RT03.}}
    \label{tab:swbd}
  % \centering
  \color{darkgray}{
  \resizebox{85mm}{19mm}{
    \begin{tabular}{c|c|c|c|c|cc|cc} 
        \toprule
           \multirow{2}{*}{\textbf{models}} & \multirow{2}{*}{\textbf{$n_{layers}$}}  & \textbf{quant.} & \multirow{2}{*}{\textbf{\#bit}} &  \multirow{2}{*}{\textbf{PPL}} & \multicolumn{4}{c}{Hub5'00 \ \  \textbf{WER(\%)}}\\ 
            & & \textbf{meth.}& & &swbd. & callhm. & rt02 & rt03 \\ \midrule
            1 & 6 & - &32  & 45.9 &7.8 & 15.6 & 12.9 & 17.3 \\ \hline
           2 & 16 &- &32  & 45.9 & 9.5 & 16.2 & 15.1 & 19.4  \\ \hline
           3 & \multirow{6}{*}{16} &  &1  & 46.2 & 9.3 & 16.8  & 15.3 &19.6\\ 
             4 & & ADMM   & 2 & 45.7 &8.1 & 16.0 & 14.7	&18.7 \\ 
             5 & & (manual) & \textbf{4} & \textbf{45.1} &\textbf{7.8} & \textbf{15.7} &\textbf{13.5}&\textbf{17.3} \\ 
              6 & &  & 8 & 45.0 & 8.0 & 16.0 & 14.0 & 18.1\\ \cline{1-1}\cline{3-9}
             7 & & MinSen & 2.8 &  45.3& 7.8 & 15.8 & 13.9 & 17.7\\ \cline{1-1}\cline{3-9}
             8 & & NAS. & 2.1 &45.6 & 7.9 & 15.9 & 14.2 & 17.9\\ \bottomrule
        \end{tabular}}}
             \vspace{-1.5em}
  \end{table}
% This paper investigates the use of alternating direction methods of multipliers (ADMM) based optimization method to directly train low-bit quantized RNNLMs from scratch. Experimental results conducted on multiple tasks suggest the proposed technique can achieve faster convergence than the baseline binarized RNNLMs quantization, while producing comparable model compression ratios. Future research will investigate the application of ADMM based quantization techniques to more advanced forms of neural language models and acoustic models for speech recognition.

\section{Acknowledgement}
This research is supported by Hong Kong RGC GRF grant No. 14200218, 14200220, TRS T45-407/19N, Innovation \& Technology Fund grant No. ITS/254/19, and SHIAE grant No. MMT-p1-19.

\bibliographystyle{IEEEbib}
\bibliography{refs}
\end{document}